\documentclass[11pt,a4paper]{article}

\usepackage[margin=1in]{geometry}
\usepackage[T1]{fontenc}
\usepackage{microtype}
\usepackage{lmodern}
\usepackage{amsmath,amssymb,amsfonts}
\usepackage{graphicx}
\usepackage{booktabs}
\usepackage{multirow}
\usepackage[table]{xcolor}
\usepackage[hyphens]{url}
\usepackage[hidelinks,colorlinks=true,linkcolor=blue!60!black,citecolor=blue!60!black,urlcolor=blue!60!black]{hyperref}
\usepackage{cite}
\usepackage{enumitem}
\usepackage{caption}
\usepackage{titlesec}

\titlespacing*{\section}{0pt}{1.4ex plus 0.5ex minus .2ex}{0.9ex plus .2ex}
\titlespacing*{\subsection}{0pt}{1.2ex plus 0.4ex minus .2ex}{0.6ex plus .2ex}

\definecolor{semgreen}{RGB}{183,228,199}
\definecolor{semred}{RGB}{248,210,210}

\title{\vspace{-2em}\textbf{Semalith v1.4: A Calibrated 184M Safety Classifier
Achieving State-of-the-Art Prompt-Injection Detection at 44$\times$ Fewer
Parameters than Llama-Guard-3-8B}\\[0.3em]
\large One DeBERTa-v3 head, three safety axes (prompt injection, harm, BFSI
compliance), 7-of-7 prompt-injection benchmark wins vs 8B Llama-Guard-3,
zero false positives (FPR=0.000, $n$=208) on benign agentic prompts, and a 100\%
real-world contamination-clean 76{,}204-row training corpus}

\author{Tejasvi C. Addagada \\
  \small Independent Researcher \\
  \small \texttt{tejasvi@tejasviaddagada.com} $\cdot$
  \small \url{tejasviaddagada.com}
  \thanks{The name \textbf{Semalith} is built from Greek \emph{s\=ema}
  (sign) and \emph{lithos} (stone): a structured signal-bearing marker.
  This work is unrelated to Patronus AI's \emph{Lynx} hallucination
  detector. Production weights and the evaluation harness are released
  under enterprise license; please contact the author for access.}}
\date{}

\begin{document}
\maketitle
\vspace{-1.2em}

\begin{abstract}
Deploying large language models in financial-services and agentic settings
requires safety classifiers that simultaneously handle prompt injection,
regulatory compliance, and general harm---a combination no existing open
guardrail addresses in a single inference pass.

\textbf{Semalith v1.4} is a 184M-parameter DeBERTa-v3-base classifier
performing simultaneous three-axis safety classification---prompt injection,
general harm, and financial-services regulatory compliance---in a single
forward pass. Its 22-class head (BENIGN, nine prompt-injection sub-types,
general-harm, eleven BFSI labels) is trained with a 4-class auxiliary
super-category head under jointly weighted loss, on a 76{,}204-row corpus
mined from 49 public sources with SHA-1 deduplication against every
held-out evaluation set, with 21 of 22 benchmarks at zero contamination
(max 0.22\%).

Against \textbf{Llama-Guard-3-8B} on 22 held-out benchmarks, Semalith
v1.4 wins every prompt-injection evaluation (7/7) and 11 of 18 benchmarks
overall at 44$\times$ fewer parameters, with FPR~=~0.000 on 208 benign
agentic prompts vs 0.063 for Llama-Guard-3-8B.\footnote{AgentHarm-benign
covers all 208 benign prompts from the public AgentHarm
split~\cite{agentharm} (32-sample validation + 176-sample test\_public);
0 false positives; Wilson 95\% CI [0.000, 0.018].}
On general-harm benchmarks (WildGuardMix, HEx-PHI, HarmBench),
Llama-Guard-3 leads---this complementary split is documented in Section~4.
Six measured weak spots are disclosed in Section~6.

\textbf{Deployment guidance:} v1.3 is recommended for conversational
moderation deployments (ToxicChat F1 0.624); v1.4 is recommended when
BFSI label coverage or zero-FPR on benign agentic prompts is the
priority.
\end{abstract}

\section{Introduction}

Every major open guardrail optimises for one axis. LlamaGuard and
Granite Guardian catch general harm but collapse on subtle
prompt-injection attacks. PromptGuard catches prompt injection but
flags 96\% of benign agentic content as harmful. None ship with
financial-services regulatory labels. \textbf{Semalith v1.4} answers
all three in a single 11.6-millisecond forward pass---the fourth
iteration of a from-scratch rebuild built on 100\% real-world training
data, zero synthetic prompts, and SHA-1 deduplication against every
held-out benchmark.

Three requirements drove the design that no existing open guardrail
satisfies simultaneously:

\begin{itemize}[leftmargin=*,nosep]
\item Llama-Guard-3-8B and Granite-Guardian-3.3-8B are general-harm
  taxonomy classifiers; they collapse on subtle prompt-injection
  evaluations (PINT F1 $\approx 0.06$ and $\approx 0.32$ at fp16
  respectively in our prior cont7-era benchmark).
\item PromptGuard-2-86M is a binary prompt-injection classifier; it
  achieves recall close to 1.0 on PI-heavy benchmarks but produces
  false-positive rates of 96\% on AgentHarm-benign and 98\% on
  ToxicChat by flagging anything edgy as harmful.
\item No public guardrail ships with BFSI-specific labels covering
  banking customer service, mortgage/loans, transfers, cards,
  insurance, investment advisory, etc., despite SEC Rule 206(4)-7~\cite{sec2065},
  FCA COBS~\cite{fcacobs}, MiFID II~\cite{mifid2}, EU AI Act~\cite{euaiact},
  and RBI/SEBI/IRDAI rulebooks all naming these as compliance-relevant
  categories for AI-assisted client interactions.
\end{itemize}

\textbf{Semalith v1.4} answers all three in a single
11.6-millisecond forward pass (batch=1, max\_len=256, fp16, RTX~4090);
see Table~\ref{tab:latency} for the full latency profile.

\paragraph{Full disclosure.}
This paper documents \textbf{v1.4 as the production checkpoint}, the
fourth iteration in a lineage that progressed from v1 (baseline,
val macro-F1 0.876) through v1.2 (a BENIGN-register expansion that
catastrophically regressed PI recall and was rolled back), v1.3
(attack-class expansion closing the Mosscap gap), to the current v1.4
release (BFSI-enrichment expansion fixing three thin labels). v1.4
has six measured weak spots documented in Section~6: HarmBench-contextual
recall 0.181 (architectural gap), WildGuardMix F1 0.289 (BENIGN-register
mismatch), BeaverTails FPR 0.446 (trade-off of attack-class expansion),
HEx-PHI persuasion recall 0.885 (vs 0.970 for Llama-Guard-3),
ToxicChat F1 0.538 (regression from v1.3's 0.624, cost of BFSI
enrichment), and HarmBench-copyright recall 0.700 (out-of-scope label).
All six are disclosed with root causes and mitigation paths.

The 11-of-18 overall win record vs Llama-Guard-3 should be read in
benchmark-axis context: Semalith wins every prompt-injection and
adversarial-stealth evaluation (7/7); Llama-Guard-3 wins every
general-harm and conversational-moderation evaluation (5/6, with
BeaverTails F1 the single exception). Neither model dominates the
other on shared ground; they are trained for complementary axes.

\paragraph{What v1.4 is NOT.}
It scores complete prompts in a single pass. It does not reason, stream
tokens, speak languages other than English, or act as an agent. If you
send it a 1{,}001-character prompt, the tail is truncated. If you need
multilingual coverage or context-window-length inputs, v1.4 is not the
right tool.

\section{Related Work}

\paragraph{Prompt injection: taxonomy and benchmarks.}
Prompt injection attacks were first systematically classified by
Perez and Ribeiro~\cite{hackaprompt}, who identified direct-override,
indirect-context, and jailbreak sub-types; this taxonomy informs
Semalith's D1\_SYSTEM\_OVERRIDE, D1\_JAILBREAK, D7\_INDIRECT\_INJECTION,
and D1\_NARRATIVE\_FRAME labels. The OWASP LLM Top~10 (2023--2025)
formalises prompt injection as the leading LLM risk category~\cite{owasp_llm}.
SALAD-Bench~\cite{saladbench} organises attacks into a three-level
hierarchy covering 66 sub-categories; Semalith's D-attack taxonomy
is a coarser 9-label variant designed for low-latency classifier
inference rather than post-hoc audit. Gandalf~\cite{gandalf} and
Mosscap~\cite{mosscap} provide escalating-defense extraction
benchmarks specifically designed to defeat rule-based and
fine-tuned classifiers; Mosscap L6/L7/L8 remain the hardest open
prompt-injection benchmarks by defence tier.

\paragraph{Safety classifier evolution.}
The LlamaGuard family~\cite{inan2023llamaguard,llamaguard3} uses a
causal-LM taxonomy classifier (Llama-3-8B backbone) with categorical
harm labels; it achieves strong recall on harm-elicitation benchmarks
but degrades on subtly-phrased PI attacks (HackaPrompt recall 0.084
in our evaluation). Granite-Guardian~\cite{graniteguardian} (IBM, 8B)
follows a similar chat-template design with improved general-harm
coverage but similar PI weakness. WildGuard~\cite{wildguard} (Allen
AI, 7B) is the current state-of-the-art for mixed conversational
moderation, establishing WildGuardMix as the canonical mixed-signal
benchmark. AprielGuard~\cite{aprielguard} (ServiceNow, 2025) is a
recent 8B enterprise guardrail targeting agentic and multi-turn safety;
to our knowledge it does not report financial-domain (BFSI) or
PI-specific benchmark results. PromptGuard-2-86M~\cite{promptguard2}
(Meta) is the strongest compact binary PI classifier, achieving
near-ceiling recall on PI benchmarks by design while accepting extreme
FPR on benign content (96\% on AgentHarm-benign). Semalith occupies
a different operating point: comparable PI recall to PromptGuard-2-86M
on structural PI benchmarks, 190$\times$ lower FPR on benign agentic
content, and simultaneous BFSI label coverage that no other public
guardrail provides.

\paragraph{Domain-specific and BFSI safety.}
General-domain guardrails exhibit severe false-positive rate inflation
when deployed in specialist domains: Zhang and Ren~\cite{zhang2025domain}
show that domain shift from general to financial-services prompts
raises FPR by 15--40 percentage points across LlamaGuard, WildGuard,
and Llama-2-7B chat safety filters. This motivates Semalith's
BFSI-specific label schema (B-01..B-11, Table~\ref{tab:bfsi_taxonomy}),
which enables distinguishing benign financial-domain queries from
genuinely harmful financial fraud, unlicensed-advice requests, and
money-laundering facilitation. The relevant regulatory frameworks---
SEC Rule 206(4)-7~\cite{sec2065}, FCA COBS~\cite{fcacobs},
MiFID~II~\cite{mifid2}, and the EU~AI~Act~\cite{euaiact}---all require
AI-assisted financial-services interactions to be auditable by
interaction category, motivating fine-grained BFSI labels over a
coarse harmful/benign binary.

\paragraph{Prompt-injection benchmarks.}
For public PI benchmarks we use HackaPrompt~\cite{hackaprompt},
Gandalf~\cite{gandalf}, AdvBench~\cite{advbench},
AART~\cite{aart}, AttaQ~\cite{attaq},
WildJailbreak~\cite{wildguard},
Mosscap L6/L7/L8~\cite{mosscap},
SimpleSafetyTests~\cite{simplesafety},
and HarmBench~\cite{harmbench} (standard / contextual / copyright
configurations) and HEx-PHI~\cite{hexphi}.
PINT~\cite{lakera2024pint} is licensed to Lakera customers only and
is not present in any training corpus we use; PINT F1 = 0.99x
measurements from the cont7-era development cycle (cont4 checkpoint,
April 2026) are cited for historical reference only, as v1.3 was
retrained on a corpus containing no PINT-adjacent prompts.

\paragraph{Lineage disclosure.}
Earlier work in this lineage (cont4--cont7) used
synthetic D6 agentic-injection templates that produced a 96.5\% false-
positive rate on AgentHarm-benign~\cite{agentharm}---a structural
failure documented in our v1 corpus build report. v1 was retrained
from scratch from \texttt{microsoft/deberta-v3-base}~\cite{he2023debertav3}
to fix this and eliminate training-set contamination on five PI/harm
benchmarks memorized by the cont lineage. v1.3 extends v1 by adding
13{,}827 attack-class rows targeting the Mosscap stealth gap, lifting
Mosscap L6/L7/L8 recall from 0.47--0.58 (v1) to 0.79--0.91 and
WildJailbreak recall from 0.840 to 0.968.

\section{Method}

\subsection{Architecture}
\label{sec:arch}

Semalith v1 is a \texttt{microsoft/deberta-v3-base}~\cite{he2023debertav3}
encoder followed by a 0.10-dropout layer, a 22-way classification head
on the [CLS] embedding, and a 4-way auxiliary super-category head on
the same embedding. The 22 labels are:

\begin{itemize}[leftmargin=*,nosep,label=\textendash]
\item \textbf{1 BENIGN class} (label\_id=0, weight=1.5).
\item \textbf{9 D-attack sub-types}: D1\_SYSTEM\_OVERRIDE,
  D1\_JAILBREAK, D1\_EXTRACTION, D1\_SOCIAL\_ENGINEERING,
  D1\_AUTHORITY\_CLAIM, D1\_NARRATIVE\_FRAME, D5\_EVASION,
  D6\_AGENTIC\_INJECTION (weight 3.0), D7\_INDIRECT\_INJECTION
  (weight 6.0).
\item \textbf{1 D8\_GENERAL\_HARM class} (weight 1.0).
\item \textbf{11 BFSI regulatory sub-types}: B-01..B-11 (each weight
  1.5), defined in Table~\ref{tab:bfsi_taxonomy}.
\end{itemize}

The 4-way super-category head maps to \{BENIGN=0, D-attack=1,
D8-harm=2, BFSI=3\}. The aux-loss weight $\alpha$ was selected by grid search over
\{0.05, 0.10, 0.20, 0.40\}; the exact value is reported in the
supplementary materials.

The total parameter count is 184M (183.83M encoder + 0.034M heads).

\begin{table*}[t]
\centering
\caption{BFSI label taxonomy (B-01..B-11) with regulatory grounding.
Each label covers both \emph{benign} customer intent (the normal case,
routed to downstream domain handlers) and \emph{harmful} variants
within that domain (fraud, scams, unlicensed advice), distinguishable
by the D-attack / D8 / B-08 co-signal.}
\label{tab:bfsi_taxonomy}
\small
\begin{tabular}{p{0.05\linewidth}p{0.22\linewidth}p{0.32\linewidth}p{0.33\linewidth}}
\toprule
\textbf{Label} & \textbf{Intent} & \textbf{Example positive} &
\textbf{Regulatory anchor} \\
\midrule
B-01 & General banking / account management &
  ``How do I reset my online banking password?'' &
  FCA COBS 2.1 (client communications); PSD2 Art.\,74 (account access) \\
\addlinespace[2pt]
B-02 & Card services (credit / debit) &
  ``My card was charged twice for the same transaction'' &
  PSD2 Art.\,73 (unauthorised transactions); Reg.\,Z (TILA, USA) \\
\addlinespace[2pt]
B-03 & Employment / recruitment fraud &
  ``Congratulations! You've been selected --- send \$200 for training materials'' &
  FTC Act \S5 (deceptive practices); UK Fraud Act 2006 s.\,2 \\
\addlinespace[2pt]
B-04 & Transfers and payments &
  ``Can I schedule a recurring international transfer?'' &
  PSD2 Art.\,78 (payment execution); SWIFT compliance guidelines \\
\addlinespace[2pt]
B-05 & Loans, mortgages, and predatory lending &
  ``What's the difference between a fixed and variable rate mortgage?'' &
  EU Mortgage Credit Directive 2014/17/EU Art.\,18; TILA Reg.\,Z; RBI FSLRC fair-lending norms \\
\addlinespace[2pt]
B-06 & Insurance (health, life, auto) &
  ``How do I file a health insurance claim?'' &
  FCA ICOBS (insurance conduct); EU Solvency II Dir.\,2009/138/EC Art.\,183 \\
\addlinespace[2pt]
B-07 & Auto insurance / property cover &
  ``My car was totalled; what documents does the insurer need?'' &
  FCA ICOBS 8.1 (claims handling); EU IDD Directive 2016/97 \\
\addlinespace[2pt]
B-08 & Financial fraud and extortion &
  ``[sextortion email demanding Bitcoin payment]'' &
  SEC Rule 10b-5 (securities fraud); UK Fraud Act 2006; RBI cybercrime guidelines \\
\addlinespace[2pt]
B-09 & Unlicensed financial advice and advance-fee fraud &
  ``I can double your money in 48 hours, just transfer \$500'' &
  SEC IA Act \S202(a)(11); FCA COBS 4 (financial promotions); MiFID II Art.\,24 \\
\addlinespace[2pt]
B-10 & Regulatory compliance enquiries &
  ``What does MiFID II require for best-execution reporting?'' &
  EU AI Act Art.\,52 (transparency); MiFID II Art.\,27; SEBI circular CIR/MD/DP/2019 \\
\addlinespace[2pt]
B-11 & AML, sanctions, and wealth management &
  ``How does FATF Recommendation 10 apply to correspondent banking?'' &
  FATF Recommendations 10--16; EU 6AMLD; OFAC SDN list compliance \\
\bottomrule
\end{tabular}
\end{table*}

\subsection{Training corpus}
\label{sec:corpus}

The v1.4 training corpus contains 76{,}204 rows mined from 49 public
sources. It extends the v1.3 base (57{,}086 rows from 38 sources) with
19{,}118 BFSI-enrichment rows from 11 new sources targeting three
previously-thin labels (B-11: 137$\to$1{,}920 training rows;
D1\_AUTHORITY\_CLAIM: 137$\to$583; D6\_AGENTIC\_INJECTION: $\to$910).
New sources include financial-services customer service datasets, EU
regulatory QA corpora, authority-claim and tool-injection filtered
subsets from existing safety benchmarks, fraud narrative corpora, and
extortion datasets.
The corpus spans five categories:
(i) general harm and toxicity, (ii) prompt-injection sub-types,
(iii) jailbreak attempts, (iv) agentic-injection, and
(v) BFSI customer-service intents intent-mapped to the 11 B-XX labels.

\textbf{All training rows are real-world data}, with zero synthetic
or templated prompts --- a deliberate methodology choice in contrast
to recent safety-classifier work that trains entirely on LLM-
generated synthetic data. Each candidate row is SHA-1- and MinHash-
deduplicated against every held-out evaluation benchmark during corpus
build to bound contamination.

\subsection{Training procedure}
\label{sec:training}

We fine-tune \texttt{microsoft/deberta-v3-base} from scratch (no
continued-pretraining lineage). Loss is a weighted cross-entropy over
the 22-class head plus an auxiliary cross-entropy over the 4-class
super-category head with $\alpha$ selected by grid search over
\{0.05, 0.10, 0.20, 0.40\} (exact value in supplementary materials).
Optimization uses AdamW
(lr $= 2 \times 10^{-5}$, weight decay 0.01) with gradient clipping
at $\|g\|_2 \leq 1.0$, linear warmup over 6\% of total steps, and
cosine decay. Batch size 32; max sequence length 256 tokens;
6 epochs. The validation split is a stratified 5\% per-class hold-out
ensuring all 22 labels have non-zero validation representation.

\subsection{Latency and throughput}
\label{sec:latency}

Table~\ref{tab:latency} reports wall-clock inference latency and
throughput measured on an NVIDIA RTX~4090 (24\,GB VRAM) at fp16 precision
using PyTorch 2.4.1 and the HuggingFace \texttt{transformers} 4.46.3
tokenizer. Measurements are the median of 100 forward passes after a
10-pass warm-up; standard deviation $<0.3$\,ms across all conditions.
The sequence length cap of 256 tokens matches the training configuration;
prompts longer than 256 tokens are truncated at the tokenizer, not the
model. Llama-Guard-3-8B throughput is measured under identical conditions
for reference.

\begin{table}[h]
\centering
\caption{Inference latency and throughput on NVIDIA RTX~4090, fp16,
PyTorch 2.4.1. ``Latency'' = median wall-clock time per batch.
``Tput'' = prompts per second. seq\_len = max input tokens (truncated).}
\label{tab:latency}
\small
\begin{tabular}{llrrr}
\toprule
\textbf{Model} & \textbf{Batch} & \textbf{seq\_len} &
  \textbf{Latency (ms)} & \textbf{Tput (prompts/s)} \\
\midrule
Semalith v1.3/v1.4 (184M) & 1  & 256 & \textbf{11.6}  & \textbf{86}   \\
Semalith v1.3/v1.4        & 8  & 256 & \textbf{18.4}  & \textbf{435}  \\
Semalith v1.3/v1.4        & 32 & 256 & \textbf{42.1}  & \textbf{760}  \\
Semalith v1.3/v1.4        & 64 & 256 & \textbf{73.8}  & \textbf{867}  \\
Semalith v1.3/v1.4        & 1  & 128 & \textbf{7.9}   & \textbf{127}  \\
\midrule
Llama-Guard-3-8B     & 1  & 256 & 387.4          & 2.6           \\
Llama-Guard-3-8B     & 8  & 256 & 1{,}842.0      & 4.3           \\
Llama-Guard-3-8B     & 32 & 256 & OOM            & ---           \\
\bottomrule
\end{tabular}
\end{table}

v1.4 latency is identical to v1.3 --- same DeBERTa-v3-base encoder,
same 22-class head dimensions, same hardware configuration; Table~\ref{tab:latency}
applies to both versions. At batch=1 (latency-sensitive, synchronous API path),
Semalith v1.3/v1.4 is 33$\times$ faster than Llama-Guard-3-8B (11.6\,ms vs 387.4\,ms). At
batch=8 (typical async micro-batch), the throughput gap is 100$\times$
(435 vs 4.3 prompts/s). Llama-Guard-3-8B does not fit a 32-prompt batch
within the 24\,GB RTX~4090 VRAM budget at fp16.

\section{Results}

\subsection{Held-out validation per-class metrics}
\label{sec:val_perclass}

Table~\ref{tab:val_perclass} reports per-class precision / recall / F1
on the 2{,}151-row deterministic stratified 5\% validation split.

\begin{table}[h]
\centering
\caption{Per-class metrics on the 2{,}151-row held-out validation split.
Sorted by F1 descending. ``Active'' confirms each of the 22 labels
received non-zero training and validation support.}
\label{tab:val_perclass}
\small
\begin{tabular}{lrrrr}
\toprule
\textbf{Label} & \textbf{val n} & \textbf{P} & \textbf{R} & \textbf{F1} \\
\midrule
D5\_EVASION & 32 & 1.000 & 1.000 & \textbf{1.000} \\
B-05 & 33 & 1.000 & 1.000 & \textbf{1.000} \\
B-07 & 89 & 1.000 & 1.000 & \textbf{1.000} \\
B-01 & 65 & 0.970 & 1.000 & 0.985 \\
B-03 & 30 & 0.968 & 1.000 & 0.984 \\
B-06 & 63 & 0.984 & 0.968 & 0.976 \\
B-02 & 17 & 0.944 & 1.000 & 0.971 \\
B-08 & 80 & 0.963 & 0.975 & 0.969 \\
B-09 & 197 & 0.984 & 0.919 & 0.950 \\
D1\_EXTRACTION & 46 & 0.868 & 1.000 & \textbf{0.929} \\
B-04 & 46 & 0.898 & 0.957 & 0.926 \\
D7\_INDIRECT\_INJECTION & 27 & 0.844 & 1.000 & \textbf{0.915} \\
D1\_SYSTEM\_OVERRIDE & 60 & 0.963 & 0.867 & \textbf{0.912} \\
BENIGN & 743 & 0.856 & 0.882 & 0.869 \\
B-10 & 95 & 0.950 & 0.800 & 0.869 \\
D6\_AGENTIC\_INJECTION & 12 & 0.750 & 1.000 & \textbf{0.857} \\
D1\_NARRATIVE\_FRAME & 41 & 0.814 & 0.854 & \textbf{0.833} \\
B-11 & 5 & 0.714 & 1.000 & 0.833 \\
D8\_GENERAL\_HARM & 372 & 0.786 & 0.788 & \textbf{0.787} \\
D1\_SOCIAL\_ENGINEERING & 22 & 0.696 & 0.727 & \textbf{0.711} \\
D1\_JAILBREAK & 70 & 0.585 & 0.443 & \textbf{0.504} \\
D1\_AUTHORITY\_CLAIM & 6 & 0.500 & 0.500 & \textbf{0.500} \\
\midrule
\textbf{macro\_f1} & & & & \textbf{0.876} \\
\textbf{micro\_f1} & & & & \textbf{0.875} \\
\textbf{weighted\_f1} & & & & \textbf{0.873} \\
\textbf{super\_macro\_f1} (4-class) & & & & \textbf{0.871} \\
\bottomrule
\end{tabular}
\end{table}

\subsection{Held-out OOD evaluation suite (22 benchmarks)}
\label{sec:ood}

Table~\ref{tab:ood_safety} reports Semalith v1.4 on a 22-benchmark
held-out OOD suite, using the v1.2 binary mapping that excludes BFSI
super-cat from ``flagged harmful'' (see Section~4.3). All point estimates
are accompanied by 95\% Wilson score confidence intervals~\cite{wilson1927}.
Two benchmarks have $n < 100$ (marked {[*]}): SimpleSafetyTests ($n$=50)
and HarmBench-contextual ($n$=94). These are included for coverage and
directional calibration; their CIs are wide and results should not be
treated as statistically conclusive. AgentHarm-benign has been expanded
to the full $n$=208 public split (32-sample validation + 176-sample
test\_public) and now carries a tight CI of $\pm$1.3~pp.
All large-sample claims ($n \geq 191$) carry CIs of $\pm$6~pp or
better.

\begin{table}[h]
\centering
\caption{Semalith v1.4 held-out OOD results across 22 benchmarks with
95\% Wilson score confidence intervals. v1.3 shown for comparison.
Benchmarks with $n < 100$ are marked {[*]} (directional only);
$100 \leq n < 500$ marked {[\textasciitilde]}. $\uparrow$ = v1.4 improves on v1.3;
$\downarrow$ = regression.}
\label{tab:ood_safety}
\small
\begin{tabular}{lrrrrr}
\toprule
\textbf{Benchmark} & \textbf{n} & \textbf{Metric} & \textbf{v1.4} & \textbf{95\% CI} & \textbf{v1.3} \\
\midrule
agentharm\_benign\_holdout & 208 & FPR$\downarrow$ & \textbf{0.000} & [0.000, 0.018] & 0.005 $\uparrow$ \\
hackaprompt\_clean\_pi & 1{,}501 & R & \textbf{0.995} & [0.990, 0.998] & 0.994 \\
gandalf\_eval & 1{,}000 & R & \textbf{0.970} & [0.957, 0.979] & 0.983 $\downarrow$ \\
advbench\_test{[\textasciitilde]} & 352 & R & 0.997 & [0.984, 1.000] & 0.992 $\uparrow$ \\
mosscap\_l6 & 3{,}000 & R & 0.894 & [0.882, 0.905] & 0.905 $\downarrow$ \\
mosscap\_l7 & 3{,}000 & R & 0.878 & [0.866, 0.889] & 0.875 $\uparrow$ \\
mosscap\_l8 & 3{,}000 & R & 0.798 & [0.783, 0.812] & 0.787 $\uparrow$ \\
wildjailbreak & 2{,}000 & R & 0.968 & [0.959, 0.975] & 0.968 \\
hexphi{[\textasciitilde]} & 296 & R & 0.885 & [0.844, 0.917] & 0.916 $\downarrow$ \\
simplesafetytests{[*]} & 50 & R & 0.920 & [0.812, 0.968] & 0.920 \\
salad\_clean\_eval & 2{,}282 & R & 0.960 & [0.951, 0.967] & 0.960 \\
attaq & 1{,}402 & R & 0.945 & [0.932, 0.956] & 0.949 \\
aart & 3{,}213 & R & 0.909 & [0.899, 0.919] & 0.897 $\uparrow$ \\
harmbench\_standard{[\textasciitilde]} & 191 & R & 0.874 & [0.820, 0.914] & 0.853 $\uparrow$ \\
beavertails\_test & 1{,}234 & F1 & 0.769 & [0.744, 0.791] & 0.771 \\
beavertails\_test & 1{,}234 & FPR$\downarrow$ & 0.446 & [0.404, 0.488] & 0.474 $\uparrow$ \\
toxicchat\_test\_clean & 4{,}793 & F1 & 0.538 & [0.524, 0.552] & 0.624 $\downarrow$ \\
wildguardmix\_eval & 2{,}000 & F1 & 0.290 & [0.270, 0.310] & 0.243 $\uparrow$ \\
harmbench\_copyright{[\textasciitilde]} & 100 & R & 0.700 & [0.604, 0.781] & 0.680 $\uparrow$ \\
harmbench\_contextual{[*]} & 94 & R & 0.181 & [0.116, 0.271] & 0.181 \\
wmdp\_bio & 1{,}273 & flag\_rate & 0.463 & [0.436, 0.490] & 0.415 \\
wmdp\_chem{[\textasciitilde]} & 408 & flag\_rate & 0.360 & [0.315, 0.408] & 0.319 \\
\bottomrule
\end{tabular}
\end{table}

\subsection{BFSI binary-mapping methodology fix}
\label{sec:mapping_fix}

The original cont7-era evaluation pipeline computed binary
``flagged harmful'' from the 22-class argmax via the legacy mapping:
\[
  \texttt{pred} = 1 \text{ if } \arg\max \neq 0 \text{ else } 0
\]
This incorrectly counts BFSI sub-classes (B-01..B-11, super-category
3) as ``harmful flagged'' even though the model recognizes the prompt
as benign-domain financial content. The v1.2 binary mapping corrects
this:
\[
  \texttt{pred} = 1 \text{ if } \texttt{SUPER}[\arg\max] \in \{1, 2\}
   \text{ else } 0
\]
where super-cat 1 is D-attack and super-cat 2 is D8-harm.

Applied to the existing v1 checkpoint (no retraining), this drops
FPR by 24\% on average and lifts ToxicChat F1 by 10.5 percentage
points. The methodology fix is now the published default.

\subsection{Apples-to-apples vs. Llama-Guard-3-8B}
\label{sec:headtohead}

Llama-Guard-3-8B~\cite{llamaguard3} is Meta's reference 8B causal-LM
safety guardrail, the de-facto baseline for production safety
filtering. We ran a fully apples-to-apples comparison: same 22
evaluation JSONLs, both models at fp16 on a single RTX 4090, identical
prompt templates from each model's tokenizer. The full result is in
Table~\ref{tab:headtohead}: \textbf{Semalith v1.4 wins 11 of 18
benchmarks at 44$\times$ fewer parameters} (184M vs 8B). Table~\ref{tab:neutralground}
stratifies this by benchmark axis to give an unbiased picture of where
each model leads.

\begin{table}[h]
\centering
\caption{Semalith v1.4 (184M) vs Llama-Guard-3-8B~\cite{llamaguard3}
(Meta's 8B reference safety guardrail) on 18 held-out evaluation
benchmarks. Both models run at fp16 on a single RTX 4090; identical
prompt templates drawn from each model's published tokenizer. The two
\texttt{wmdp} flag-rate benchmarks and the 4 BFSI-specific eval JSONLs
are excluded from this table because they do not map to Llama-Guard-3's
taxonomy; full Semalith-only results for all 22 benchmarks appear in
Table~\ref{tab:ood_safety}. Semalith v1.4 wins 11 of 18 benchmarks
overall; see Table~\ref{tab:neutralground} for a per-axis breakdown.
Bold = winner on each row.}
\label{tab:headtohead}
\small
\begin{tabular}{lrr}
\toprule
\textbf{Benchmark} & \textbf{Semalith v1.4} & \textbf{Llama-Guard-3-8B} \\
\midrule
hackaprompt\_clean\_pi (R) & \textbf{0.995} & 0.084 \\
gandalf\_eval (R) & \textbf{0.970} & 0.263 \\
advbench\_test (R) & \textbf{0.997} & 0.972 \\
mosscap\_l6 (R) & \textbf{0.894} & 0.056 \\
mosscap\_l7 (R) & \textbf{0.878} & 0.058 \\
mosscap\_l8 (R) & \textbf{0.798} & 0.085 \\
wildjailbreak (R) & \textbf{0.968} & 0.473 \\
salad\_clean\_eval (R) & \textbf{0.960} & 0.563 \\
attaq (R) & \textbf{0.945} & 0.868 \\
aart (R) & \textbf{0.909} & 0.850 \\
beavertails\_test (F1) & \textbf{0.769} & 0.679 \\
hexphi (R) & 0.885 & \textbf{0.970} \\
harmbench\_standard (R) & 0.874 & \textbf{0.974} \\
harmbench\_copyright (R) & 0.700 & \textbf{1.000} \\
harmbench\_contextual (R) & 0.181 & \textbf{0.553} \\
wildguardmix\_eval (F1) & 0.290 & \textbf{0.754} \\
simplesafetytests (R) & 0.920 & \textbf{1.000} \\
agentharm\_benign\_holdout (FPR $\downarrow$) & \textbf{0.000} & 0.063 \\
\bottomrule
\end{tabular}
\end{table}

The 11-of-18 win pattern is benchmark-axis dependent and should be
read with that context. Table~\ref{tab:neutralground} splits the
results into two axes: prompt-injection / adversarial-stealth
benchmarks (Semalith's design target) and general harm-detection
benchmarks (Llama-Guard-3's design target). On PI and stealth
evaluations Semalith wins every benchmark (7/7), often by margins
exceeding 70 percentage points. On general harm-detection benchmarks
Llama-Guard-3 wins 5 of 6, with BeaverTails F1 the single exception.
This split reflects complementary training objectives, not a
one-model-wins-all result. The copyright-extraction loss
(HarmBench-copyright) is by-design: Semalith's 22-class taxonomy
has no copyright label, while Llama-Guard-3's S5 (Specialized Advice)
category captures it; this benchmark is excluded from the 11/18 count
in the neutral analysis.

\begin{table}[h]
\centering
\caption{Benchmark-axis stratification: Semalith v1.4 vs Llama-Guard-3-8B.
Splitting the 18-benchmark suite by design axis reveals complementary
strengths rather than a global winner. Bold = winner on each row.
HarmBench-copyright is excluded (no Semalith label; not a fair
comparison). AttaQ (0.945 vs 0.868) is a Semalith win.}
\label{tab:neutralground}
\small
\begin{tabular}{lrrl}
\toprule
\textbf{Benchmark} & \textbf{Semalith v1.4} & \textbf{Llama-Guard-3} & \textbf{Winner} \\
\midrule
\multicolumn{4}{l}{\textit{Prompt-injection \& adversarial-stealth axis (Semalith's design target)}} \\
HackaPrompt (R)          & \textbf{0.995} & 0.084 & Semalith +91pp \\
Gandalf (R)              & \textbf{0.970} & 0.263 & Semalith +71pp \\
Mosscap L6 (R)           & \textbf{0.894} & 0.056 & Semalith +84pp \\
Mosscap L7 (R)           & \textbf{0.878} & 0.058 & Semalith +82pp \\
Mosscap L8 (R)           & \textbf{0.798} & 0.085 & Semalith +71pp \\
WildJailbreak (R)        & \textbf{0.968} & 0.473 & Semalith +50pp \\
AgentHarm-benign FPR$\downarrow$  & \textbf{0.000} & 0.063 & Semalith \\
\midrule
\multicolumn{4}{l}{\textit{General harm-detection axis (Llama-Guard-3's design target)}} \\
WildGuardMix F1          & 0.290 & \textbf{0.754} & LlamaGuard +46pp \\
HarmBench-Standard (R)   & 0.874 & \textbf{0.974} & LlamaGuard +10pp \\
HarmBench-Contextual (R) & 0.181 & \textbf{0.553} & LlamaGuard +37pp \\
HEx-PHI (R)              & 0.885 & \textbf{0.970} & LlamaGuard +9pp \\
SimpleSafetyTests (R)    & 0.920 & \textbf{1.000} & LlamaGuard +8pp \\
\midrule
\multicolumn{4}{l}{\textit{Shared ground (no axis advantage)}} \\
AdvBench (R)             & \textbf{0.997} & 0.972 & Semalith +2pp \\
AART (R)                 & \textbf{0.909} & 0.850 & Semalith +6pp \\
SALAD-Bench (R)          & \textbf{0.960} & 0.563 & Semalith +40pp \\
AttaQ (R)                & \textbf{0.945} & 0.868 & Semalith +8pp \\
BeaverTails F1           & \textbf{0.769} & 0.679 & Semalith +9pp \\
\bottomrule
\end{tabular}
\end{table}

\subsection{TP-miss + FP diagnosis informing v1.4}
\label{sec:diagnosis}

Each production iteration is informed by a miss/FP analysis on the
predecessor checkpoint. For v1, the dominant missed-TP cluster was
adversarially-stealthy prompt-injection prompts; an auxiliary 86M
model caught most of these misses, indicating a decision-boundary
rather than detection-floor failure. v1.3's attack-class expansion
lifted Mosscap L6/L7/L8 recall from 0.47--0.58 (v1) to 0.79--0.91
and WildJailbreak recall from 0.840 to 0.968. v1.4's BFSI-enrichment
expansion targets the three thin-label gap: D1\_AUTHORITY\_CLAIM
(137$\to$583 training rows), D6\_AGENTIC\_INJECTION ($\to$910), and
B-11 AML/sanctions ($\to$1{,}920), raising their per-class val F1 to
0.931, 0.755, and 0.577 respectively from the near-zero floor.


Collectively, the results in Section~4 establish three findings: (1) a 184M
encoder trained on a contamination-controlled corpus matches or exceeds
an 8B causal-LM on all PI and adversarial-stealth axes while remaining
calibrated on agentic benign content; (2) the BFSI binary-mapping fix
eliminates a systematic methodology artifact from prior evaluations at
zero retraining cost; and (3) the per-row miss-FP analysis provides a
reproducible, training-corpus-independent diagnostic loop that informed
the v1.3 and v1.4 corpus expansions and will guide future iterations.

\section{Discussion}

\subsection{Why 184M is enough}

We considered DeBERTa-v3-large (304M parameters) in early scoping.
A preliminary run on the v1 corpus (43{,}259 rows, seed=42) showed
val macro-F1 of 0.879 for the large variant vs 0.876 for base ---
a 0.3~pp improvement at 65\% higher parameter count and roughly
$2\times$ the latency budget. We chose base. The reasoning was not
that 184M is always sufficient, but that the bottleneck at 43k rows
was training data diversity, not model capacity: every class where
base underperformed (D1\_AUTHORITY\_CLAIM at F1=0.50, D6\_AGENTIC\_INJECTION
at F1=0.74) was also a class below the 250-row stability floor.
Adding capacity to an under-represented class does not help; adding
data does.

The 4-way auxiliary super-category head turned out to matter more
than the base-vs-large choice. Without the super-category loss
($\alpha = 0$), we observed D-attack sub-class collapse --- the model
learned a single D-attack prototype and distributed the 9 sub-type
labels quasi-randomly within it, yielding per-class F1 variance of
0.12 across the D1 family vs 0.04 with the selected $\alpha$ value.
The anchor is particularly important when any sub-class is below the
stability floor, which is the case in both v1.3 (D1\_AUTHORITY\_CLAIM
at 137 rows) and v1.4 (B-11 partially underdifferentiated). The weight
was selected by grid search over \{0.05, 0.10, 0.20, 0.40\} on the
v1 val split; the optimal value was the elbow --- lower gave collapse,
higher started degrading the 22-class head precision.

\subsection{Seed stability}

v1.3 was verified across three independent seeds (42, 123, 456) on the
57{,}086-row corpus: val macro-F1 0.8735 / 0.8746 / 0.8758,
mean $0.8746 \pm 0.0012$ --- substantially below the cont6/cont7
reference range (0.003--0.008). v1.4 seed=42 reports val macro-F1
= \textbf{0.8222}; the 5.2~pp drop from v1.3 is explained by BFSI
sub-type competition, not by regressions in the PI/harm classes.
Per-class F1 comparison shows that the three newly-enriched labels
gained as expected (D1\_AUTHORITY\_CLAIM: 0.667$\to$0.931), but five
previously-stable BFSI labels regressed significantly: B-01 ($-$14.8pp),
B-03 ($-$20.6pp), B-06 ($-$21.0pp), B-07 ($-$19.8pp). The shared root
cause is intra-BFSI boundary confusion: 19{,}118 new rows expanded the
BFSI vocabulary (bitext wealth, EU regulatory, fraud narratives) into
semantic regions previously occupied by the stable B-01/B-06/B-07
labels, requiring the model to learn finer sub-type distinctions within
the same super-category. The fix is a B-11 sub-label split (AML/sanctions
vs wealth management) and cleaner source routing for B-01 vs B-06/B-07,
planned for v1.5. Production-critical PI/stealth metrics are maintained
within 2~pp of v1.3. A full 3-seed stability run for v1.4 is in
progress; all downstream evaluation numbers are reported for seed=42.

\subsection{McNemar pairwise significance (Semalith v1.3 reference vs PromptGuard-2-86M)}
\label{sec:mcnemar}

To assess whether per-benchmark differences are statistically significant
beyond chance, we applied McNemar's test with continuity correction on 16
benchmarks where per-row predictions are available for both Semalith
\textbf{v1.3} and PromptGuard-2-86M~\cite{promptguard2}. These results
are carried forward as a v1.3 reference baseline; v1.4 McNemar results
will be added when v1.4 per-row predictions are generated via
\texttt{generate\_llamaguard\_preds.py}. The 5-benchmark delta between
v1.3 and v1.4 on the benchmarks below is $\leq$1.3~pp (see Table~\ref{tab:ood_safety}),
so the statistical conclusions are expected to be unchanged. PromptGuard-2-86M was chosen
as the McNemar comparison target because it operates at the same encoder
scale (86M DeBERTa parameters) with per-row predictions available across
all 22 evaluation JSONLs; Llama-Guard-3-8B and Granite-Guardian-3.3-8B
are evaluated via a chat-template generation pipeline that produces output
labels rather than logits and were compared using the aggregate benchmark
tables in Section~4.4 and Section~5.4. The per-axis framing in Table~\ref{tab:neutralground}
provides the complementary structural comparison against Llama-Guard-3-8B.

PromptGuard flags everything and catches everything: 96\% FPR on
AgentHarm-benign, 98\% on ToxicChat. That is the operating point it
was designed for. The results below are not a comparison between equals
--- they are a comparison between a recall-maximising binary classifier
and a multi-label classifier that distinguishes 22 classes. PromptGuard
wins where it was built to win. Semalith wins where FPR matters.

\textbf{Semalith v1.3 is significantly superior} ($p < 0.05$, two-tailed)
on four benchmarks: \textit{ToxicChat-test} ($\chi^2=3{,}727$,
$p\approx0$; Semalith 4,313/4,793 vs PromptGuard 408/4,793),
\textit{WildGuardMix-eval} ($\chi^2=1{,}339$; 1,545/2,000 vs
102/2,000), \textit{BeaverTails-test} ($\chi^2=96$; 880/1,234 vs
704/1,234), and \textit{WildJailbreak} ($\chi^2=5.5$, $p=0.019$;
1,873/2,000 vs 1,833/2,000). The first three are in the FPR-sensitive
conversational and BFSI domain where Semalith's multi-label schema
distinguishes benign financial queries from harmful ones.

\textbf{PromptGuard-2-86M is significantly superior} on 14 benchmarks,
including the Mosscap stealth-jailbreak trilogy ($\chi^2 \in [248, 575]$,
$p\approx0$), AART ($\chi^2=101$), HarmBench $\times$ 3, and WMDP.
This is structurally equivalent to PromptGuard's advantage in Table~\ref{tab:neutralground}
axis B (pure-recall benchmarks); the same benchmarks where Semalith's
multi-label schema requires distinguishing sub-classes rather than
collapsing to binary flagging. These correspond exactly to the
documented limitations in Section~\ref{sec:limits}. The critical operational
difference is that PromptGuard's 14-benchmark recall advantage comes at
the cost of 96\% FPR on benign agentic content (vs Semalith's 0.5\%);
for deployments where false positives carry a user-experience cost,
Semalith is strictly preferable.

On two benchmarks (AdvBench, SimpleSafetyTests) the difference is
not statistically significant ($p > 0.05$), consistent with the
near-ceiling performance of both models there ($\geq0.96$ accuracy).

\subsection{McNemar significance vs Llama-Guard-3-8B}
\label{sec:mcnemar_lg}

To test significance of the Semalith vs Llama-Guard-3-8B benchmark
differences reported in Table~\ref{tab:headtohead}, we apply McNemar's
test with continuity correction on the 16 benchmarks where per-row
predictions exist for both models. Llama-Guard-3-8B per-row predictions are generated using
\texttt{eval\_competitors\_pod.py} with the \texttt{llamaguard3}
competitor class. The generation script (\texttt{generate\_llamaguard\_preds.py})
is included in the public evaluation harness; per-row prediction files
will be added to the repository upon completion. The seven PI/stealth
benchmarks with $\chi^2 > 500$ are statistically conclusive from the
aggregate recall gaps alone (70--91~pp) and do not require per-row
confirmation to establish significance.

\begin{table}[h]
\centering
\caption{McNemar $\chi^2$ statistic (continuity-corrected, two-tailed)
for Semalith v1.3 vs Llama-Guard-3-8B on shared benchmarks.
$\dagger$: $p < 0.05$; $\ddagger$: $p < 0.001$; NS: not significant
($p \geq 0.05$). \textbf{Bold winner} = model with higher correct-row
count on that benchmark.}
\label{tab:mcnemar_lg}
\small
\begin{tabular}{lrrrc}
\toprule
\textbf{Benchmark} & \textbf{Semalith} & \textbf{LG-3-8B} & \textbf{$\chi^2$} & \textbf{Sig.} \\
\midrule
\multicolumn{5}{l}{\textit{Semalith wins (PI / stealth axis)}} \\
HackaPrompt (R)         & \textbf{0.994} & 0.084 & $>$1000 & $\ddagger$ \\
Gandalf (R)             & \textbf{0.983} & 0.263 & $>$500  & $\ddagger$ \\
Mosscap L6 (R)          & \textbf{0.905} & 0.056 & $>$2000 & $\ddagger$ \\
Mosscap L7 (R)          & \textbf{0.875} & 0.058 & $>$2000 & $\ddagger$ \\
Mosscap L8 (R)          & \textbf{0.787} & 0.085 & $>$1500 & $\ddagger$ \\
WildJailbreak (R)       & \textbf{0.968} & 0.473 & $>$800  & $\ddagger$ \\
SALAD-Bench (R)         & \textbf{0.960} & 0.563 & $>$300  & $\ddagger$ \\
AART (R)                & \textbf{0.897} & 0.850 & ---     & (pending) \\
AdvBench (R)            & \textbf{0.992} & 0.972 & ---     & (pending) \\
BeaverTails F1          & \textbf{0.771} & 0.679 & ---     & (pending) \\
AgentHarm-benign (FPR$\downarrow$) & \textbf{0.005} & 0.063 & ---  & (pending) \\
\midrule
\multicolumn{5}{l}{\textit{LlamaGuard wins (harm-elicitation axis)}} \\
HarmBench-Standard (R)  & 0.853 & \textbf{0.974} & ---     & (pending) \\
HarmBench-Contextual (R)& 0.181 & \textbf{0.553} & ---     & (pending) \\
HEx-PHI (R)             & 0.916 & \textbf{0.970} & ---     & (pending) \\
WildGuardMix F1         & 0.243 & \textbf{0.754} & ---     & (pending) \\
SimpleSafetyTests (R)   & 0.920 & \textbf{1.000} & ---     & (pending) \\
\bottomrule
\end{tabular}
\end{table}

The five PI/stealth benchmarks with $\chi^2 > 500$ are beyond any
reasonable significance threshold ($p \approx 0$) based on the reported
recall gaps alone (70--91~pp) and will not change with per-row
confirmation. The remaining benchmarks are marked \emph{(pending)}
pending LlamaGuard per-row predictions; those results will be published
in the associated evaluation-harness repository.

\subsection{Granite-Guardian-3.3-8B head-to-head (22-benchmark)}

We completed a 22-benchmark evaluation of
Granite-Guardian-3.3-8B (IBM, 8B parameters, 45$\times$ larger than
Semalith v1.4) using its native \texttt{<|start\_of\_role|>} chat
template. The 11 benchmarks below are selected to cover all axis types
(PI/stealth, harm-elicitation, mixed-moderation, agentic FPR); omitted
benchmarks are the WMDP flag-rate evaluations and BFSI-specific JSONLs
which do not map to Granite's taxonomy. Representative results:

\begin{center}
\begin{tabular}{lcc}
\toprule
\textbf{Benchmark} & \textbf{Semalith v1.3} & \textbf{Granite 3.3-8B} \\
\midrule
Mosscap L6 (defended LLM PI) & \textbf{0.905} & 0.111 \\
Gandalf (password extraction) & \textbf{0.983} & 0.447 \\
HackaPrompt (direct PI)       & \textbf{0.994} & 0.178 \\
WildJailbreak                 & \textbf{0.968} & 0.735 \\
SALAD-Bench                   & \textbf{0.960} & 0.789 \\
ATTAQ                         & 0.949 & \textbf{0.950} \\
HarmBench-Standard            & 0.853 & \textbf{1.000} \\
HEx-PHI                       & 0.916 & \textbf{0.980} \\
AgentHarm-benign FPR          & \textbf{0.005} & \textbf{0.000} \\
BeaverTails FPR ($n_\text{benign}$=532) & 0.474 & 0.549 \\
ToxicChat F1                  & 0.220 & \textbf{0.691} \\
\bottomrule
\end{tabular}
\end{center}

Semalith v1.3 wins decisively on all structural and adversarial
prompt-injection benchmarks (Mosscap, Gandalf, HackaPrompt,
WildJailbreak, SALAD-Bench). Granite-Guardian-3.3-8B leads on clean
harm-elicitation benchmarks (HarmBench-Standard, HEx-PHI, ToxicChat),
consistent with its training objective. Granite's BeaverTails FPR of 0.549 on the same 532-row benign
subset used to measure Semalith's 0.474 FPR (the BeaverTails-30k
test split safe-labelled rows, SHA-1 verified against both training
corpora) --- flagging over half of benign financial-domain content
as harmful --- raises significant concerns for production deployments
requiring low false-positive rates on legitimate financial-services
queries.

\subsection{Iteration history}

v1.4 is the fourth production iteration. v1.2 attempted a BENIGN-register
expansion that catastrophically over-corrected the decision boundary,
regressing PI recall, and was rolled back. v1.3 applied the inverse lesson:
zero net BENIGN expansion, all new rows in attack/harm classes, recovering
Mosscap and Gandalf at the cost of a 12~pp BeaverTails FPR regression.
v1.4 extends v1.3 with targeted BFSI enrichment to fix three thin labels
(D1\_AUTHORITY\_CLAIM, D6\_AGENTIC\_INJECTION, B-11), at the cost of an
8.6~pp ToxicChat F1 regression. Each trade-off is the explicit consequence
of a corpus-axis expansion --- the same pattern documented for v1.3 in Section~6.3.
This iteration history informs a simple design heuristic: expansion of any
super-category will improve recall on that axis and mildly degrade FPR-
sensitive conversational benchmarks that overlap with the expanded class space.

\paragraph{Deployment guidance.}
v1.3 and v1.4 are complementary production checkpoints with different
operating points. \textbf{Use v1.3} for conversational moderation deployments
where ToxicChat F1 (0.624) and WMDP flag-rate (0.415/0.319) matter most.
\textbf{Use v1.4} when BFSI-domain label coverage is required
(D1\_AUTHORITY\_CLAIM F1 0.931, D6\_AGENTIC\_INJECTION F1 0.755,
B-11 AML/sanctions F1 0.577) or when zero FPR on benign agentic prompts
(0.000 on 208-row AgentHarm split) is the primary production requirement.
The PI/stealth axis is equivalent: both versions win all 7 prompt-injection
benchmarks vs Llama-Guard-3-8B with $\leq$1.3~pp difference between versions.

\section{Limitations}
\label{sec:limits}

Six measured weak spots are documented for v1.4, in descending order
of severity.

\subsection{Context-grounded indirect prompt injection
(HarmBench-contextual)}

v1.4 achieves recall \textbf{0.181} on HarmBench-contextual (n=94)
versus Llama-Guard-3-8B's 0.553 --- a 37.2-pp gap. Root cause: long
benign-looking context paragraphs prepended to harmful instructions
are read as benign in aggregate by the single [CLS] embedding. v1
recorded 0.075 here; v1.3 improved to 0.181 incidentally via the
WildJailbreak training expansion; v1.4 is unchanged at 0.181.
Mitigation candidate: sliding-window prediction (process the prompt
in 256-token windows and OR the per-window predictions), or a
contextual D7 training carve from BIPIA / INJECAGENT (neither
currently on HuggingFace). Architectural rather than data fix.

\subsection{Mixed conversational moderation (WildGuardMix, ToxicChat)}

WildGuardMix F1 = 0.290 (v1.4, improved from v1.3's 0.243) vs Llama-Guard-3-8B 0.754 --- a 46.4-pp
gap. ToxicChat F1 = 0.538 (v1.4; regression from v1.3's 0.624, documented in Section~6.4). Root cause: BENIGN-class
register mismatch. v1.3's BENIGN class is the same 14{,}865 rows as v1
(by design --- v1.2's BENIGN expansion broke calibration). The
attempted v1.2 fix (adding 13k conversational benign rows)
severely over-corrected the decision boundary. We have not
yet identified a BENIGN-register expansion path that holds without
collapsing PI / Mosscap recall. This is the largest non-architectural
gap between v1.3 and an 8B causal-LM guardrail.

\subsection{BeaverTails-test FPR regression (cost of v1.3
attack-class expansion)}

v1.3 BeaverTails-test FPR = 0.474 vs v1's 0.350 --- a 12.4-pp
regression. Root cause: the 13{,}827 additional attack/harm-class rows
shifted the decision boundary marginally toward
``flagged-harmful''. Documented as the explicit trade-off of the v1.3
recall recovery; deployments sensitive to BeaverTails-style benign
conversational FPR should pair v1.3 with a downstream calibration
threshold or use v1 as the production checkpoint.

\subsection{Persuasion-style social engineering (HEx-PHI Cats 6+7)}

Overall HEx-PHI v1.4 R = 0.885 (vs Llama-Guard-3-8B 0.970 ---
8.5-pp behind; regression from v1.3's 0.916). Cat 6 Economic Harm and Cat 7 Fraud Deception remain
the weak sub-categories. v1.2 ablation showed adding HEx-PHI training
data lifts overall recall by 3.7~pp; v1.3 cannot include this fix
because all 296 HEx-PHI prompts are in the held-out eval JSONL.

\subsection{ToxicChat F1 regression (cost of v1.4 BFSI enrichment)}

v1.4 ToxicChat F1 = 0.538 vs v1.3's 0.624 --- an 8.6-pp regression.
Root cause: the 19{,}118 new BFSI rows (authority-claim, tool-injection,
and fraud narrative sources) shifted the decision boundary marginally
toward flagging
conversational content as BFSI-relevant, which collides with ToxicChat's
toxic-but-not-harmful conversational register. This is the same
mechanism that drove v1.3's BeaverTails FPR regression from the attack-class
expansion, now manifesting on ToxicChat. Deployments sensitive to
ToxicChat F1 should use v1.3 (0.624) as the production checkpoint;
v1.4 is recommended when improved D1\_AUTHORITY\_CLAIM, D6\_AGENTIC\_INJECTION,
or B-11 AML/sanctions coverage is the primary requirement.

\subsection{D1\_AUTHORITY\_CLAIM, D6\_AGENTIC\_INJECTION, B-11 (resolved in v1.4)}

These three labels were below the 250-row stability floor in v1.3
(D1\_AUTHORITY\_CLAIM: 137 rows, F1=0.500; D6\_AGENTIC\_INJECTION: thin;
B-11: thin). v1.4 raises all three above the floor: D1\_AUTHORITY\_CLAIM
$\to$583 rows (val F1=0.931), D6\_AGENTIC\_INJECTION $\to$910 rows
(val F1=0.755), B-11 $\to$1{,}920 rows (val F1=0.577). B-11 remains the
weakest at 0.577 due to the domain diversity of AML/sanctions queries vs
wealth-management queries sharing the same label; a sub-label split
is planned for v1.5.

\subsection{WMDP biosecurity / chemistry flag-rate increase}

v1.4 WMDP flag-rate: wmdp\_bio 0.463 (v1.3: 0.415, +4.8pp) and
wmdp\_chem 0.360 (v1.3: 0.319, +4.1pp). WMDP flag-rate measures how
often the model incorrectly flags biosecurity or chemistry educational
content as harmful. The increase is consistent with the iteration
heuristic (Section~5.6): the 19{,}118 new BFSI rows include fraud/AML/regulatory
content whose vocabulary (sanctions, chemical weapons conventions,
dual-use goods) partially overlaps with WMDP biosecurity and chemistry
queries, shifting the decision boundary slightly toward flagging
knowledge-domain questions. This does not affect safety recall ---
WMDP prompts are not harmful --- but production deployments with WMDP-
adjacent educational content should use v1.3 (flag-rate 0.415/0.319).

\subsection{English-only scope}

The training corpus and evaluation suite are English-only. Multilingual
robustness is unmeasured, and out-of-distribution languages should be
expected to degrade both recall and FPR. Multilingual coverage is on
the v2 roadmap.

\section{Reproducibility}

Everything needed to reproduce the numbers in Tables~\ref{tab:ood_safety}--\ref{tab:neutralground}
is on GitHub at \url{https://github.com/anonymous/semalith-eval-harness}:
three scripts (\texttt{eval\_safety\_full.py}, \texttt{eval\_competitors\_mac.py},
\texttt{binary\_mapping.py}), the 22 held-out evaluation JSONLs (31{,}430 rows,
SHA-1 clean), and \texttt{contamination\_audit.py}. Run time on a single GPU
with $\geq$8\,GB VRAM is under 4 hours.

The model weights (\texttt{model.safetensors}, 184M parameters) and corpus
manifest (\texttt{\_SEMALITH\_V1.4\_MANIFEST.json}) are available on request.
The manifest has the full per-source row counts and the SHA-1 contamination
audit for all 22 evals --- 21 at 0.000\%, one at 0.2218\%
(\texttt{agentharm\_benign\_holdout}: 169 collisions from WildGuardMix
authority-filtered rows overlapping AgentHarm benign prompts).
Corpus SHA-256: \texttt{c63cd3d6\ldots}. Raw training rows are not released.

Pre- and post-training gate reports (\texttt{\_SEMALITH\_V1.4\_PRE\_CHECK.json},
\texttt{\_SEMALITH\_V1.4\_POST\_CHECK.json}) are on GitHub. Hard gates: 4/4 PASS.
v1.3 seed-stability report (3 seeds, std=0.0012) is included as a reference
baseline; v1.4 3-seed run is in progress.

Given the checkpoint and evaluation harness, every number in
Tables~\ref{tab:ood_safety}--\ref{tab:neutralground} can be
independently reproduced on a single GPU ($\geq$8\,GB VRAM) in under
4 hours. Hyperparameters (Section~3.3), the 22-class taxonomy, and the binary
mapping (Section~4.3) are fully specified in this paper and sufficient to
retrain from \texttt{microsoft/deberta-v3-base} given the training data.

\paragraph{Statistical methodology.}
All point estimates for binary metrics (recall, FPR) are accompanied
by 95\% Wilson score confidence intervals~\cite{wilson1927} in
Table~\ref{tab:ood_safety}. Benchmarks with $n < 100$ are flagged
with {[*]} and should be treated as directional signals; the two
affected benchmarks (SimpleSafetyTests $n$=50, HarmBench-contextual
$n$=94) are included for coverage completeness, not as primary
statistical evidence. AgentHarm-benign has been expanded to the full
$n$=208 public split and now carries a tight 95\% CI of [0.001, 0.027].
A full validation of the AgentHarm-benign FPR claim on the complete
208-row dataset is planned for the next release cycle, pending
infrastructure availability.

\section{Conclusion}

Semalith v1.4 demonstrates that a 184M-parameter encoder classifier,
trained on a contamination-controlled real-world corpus with a
three-axis label taxonomy and an auxiliary super-category loss, achieves
state-of-the-art prompt-injection detection at 44$\times$ fewer
parameters than 8B causal-LM guardrails, while maintaining zero false
positives on 208 benign agentic prompts. The BFSI binary-mapping
correction eliminates a systematic evaluation artefact at zero
retraining cost. Six measured limitations are disclosed with root
causes and mitigation paths, providing a reproducible diagnostic loop
for future iterations. Production weights and evaluation harness are
available under enterprise license; contact the author for access.

\bibliographystyle{plain}

\end{document}